\pdfoutput=1
\documentclass{article}

\usepackage{arxiv}

\usepackage[utf8]{inputenc} %
\usepackage[T1]{fontenc}    %

\usepackage{amsmath,amsfonts,bm}

\def\eqref#1{equation~\ref{#1}}

\def\1{\bm{1}}

\DeclareMathAlphabet{\mathsfit}{\encodingdefault}{\sfdefault}{m}{sl}
\SetMathAlphabet{\mathsfit}{bold}{\encodingdefault}{\sfdefault}{bx}{n}

\usepackage{amssymb}
\usepackage{amsfonts}       %
\usepackage{multirow}
\usepackage{xcolor}
\usepackage{xspace}
\usepackage{enumitem}
\usepackage{float}          %
\usepackage{booktabs}       %
\usepackage{graphicx}
\usepackage{url}            %
\usepackage{microtype}      %
\usepackage{hyperref}       %
\usepackage{natbib}
\setcitestyle{authoryear,round,citesep={;},aysep={,},yysep={;}}

\newcommand{\modelname}{DreamFAR\xspace}

\title{DreamFAR: Real-Time \& Long-Horizon \\ Interaction with Diverse Objects}

\author{%
  Yiming Dou$^{1,2}$ \quad Botao He$^{1,3}$  \quad Antonio Loquercio$^{4}$ \\
  \bfseries Andrew Owens$^{2}$ \quad Jitendra Malik$^{1}$ \quad Sumedh Anand Sontakke$^{1}$ \quad Ruoshi Liu$^{1,3}$ \\[0.8ex]
  {\small $^{1}$Amazon FAR \quad $^{2}$Cornell University \quad $^{3}$University of Maryland, College Park \quad $^{4}$University of Pennsylvania}%
}

\date{}                               %
\renewcommand{\headeright}{}          %
\renewcommand{\undertitle}{}          %
\renewcommand{\shorttitle}{}                %

\makeatletter
\renewcommand{\@maketitle}{%
  \vbox{%
    \hsize\textwidth
    \linewidth\hsize
    \vskip 0.1in
    \@toptitlebar
    \centering
    {\LARGE\bfseries \@title\par}
    \@bottomtitlebar
    \def\And{%
      \end{tabular}\hfil\linebreak[0]\hfil%
      \begin{tabular}[t]{c}\bf\rule{\z@}{24\p@}\ignorespaces%
    }
    \def\AND{%
      \end{tabular}\hfil\linebreak[4]\hfil%
      \begin{tabular}[t]{c}\bf\rule{\z@}{24\p@}\ignorespaces%
    }
    \begin{tabular}[t]{c}\bf\rule{\z@}{24\p@}\@author\end{tabular}%
    \vskip 0.3in \@minus 0.1in
  }
}
\makeatother

\hypersetup{
pdftitle={DreamFAR: Real-Time and Long-Horizon Interaction with Diverse Objects},
pdfauthor={Yiming Dou, Botao He, Antonio Loquercio, Andrew Owens, Jitendra Malik, Sumedh Anand Sontakke, Ruoshi Liu},
hidelinks,
}

\begin{document}

\maketitle

\begin{figure}[H]\centering
\IfFileExists{figures/teaser.pdf}{\includegraphics[width=\linewidth]{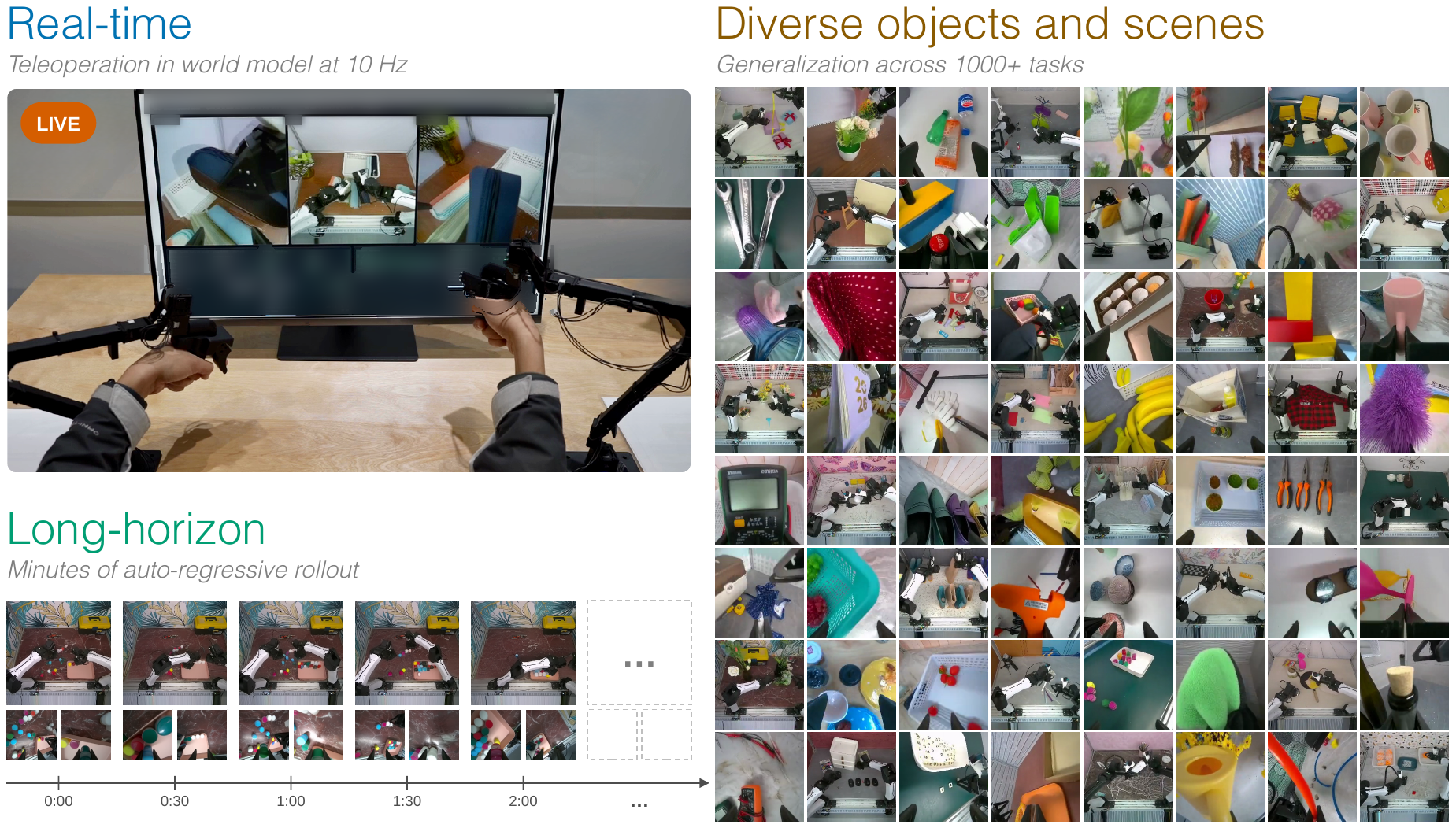}}{\fbox{\parbox[c][2.7cm][c]{0.97\linewidth}{\centering placeholder: teaser }}}
\caption{\modelname{} is a world model that is trained from a large robot-native dataset. It supports real-time interaction at 10\,Hz, persists minutes of auto-regressive rollout, and generalizes across more than 1,000 tasks. We further demonstrate through experiments that \modelname{} can be used for policy evaluation and mixed-data policy co-training. See more examples and the interactive demo on the project page: \href{https://www.yimingdou.com/DreamFAR/}{\texttt{www.yimingdou.com/DreamFAR}}.}
\label{fig:teaser}
\end{figure}

\begin{abstract}

World models have shown great potential in serving as a simulator for robot manipulation. However, existing models are either fit to a single environment or lack ability to simulate long-horizon contact-rich events. We present \modelname, a world model whose dynamics model is trained from scratch on a large robot-native teleoperation dataset, with no internet-video pretraining. We present an architecture and learning recipe designed for training on robot-native data. \modelname is significantly smaller than the high-performance video generation models, yet shows better understanding of contact and causality, and runs 10 FPS on a commercial GPU. Experimentally, \modelname significantly outperforms existing models in predicting the long-horizon future. We further demonstrate various applications where \modelname can be used, including policy evaluation and mixed-data co-training. Project page: \href{https://www.yimingdou.com/DreamFAR/}{\texttt{www.yimingdou.com/DreamFAR}}
\end{abstract}

\section{Introduction}\label{sec:intro}
A world model~\citep{ha2018world} predicts the future conditioned on an action input. The major appeal of learning such model is that it's a powerful abstraction of the world for dynamics and control, allowing robots to plan many steps into the future, simulate the outcomes of their behavior and like humans~\citep{flanagan1997role}, search for the optimal actions in order to achieve their goals. \emph{Given these strengths, what prevents world models from being the prominent tool for solving contact-rich manipulation tasks?}

One of the major shortcomings of today's world models is their inability to do {\em long-horizon prediction}~\citep{gao2026dreamdojo,ye2026dreamzero,ali2025cosmospredict}. Rolling out future predictions for long horizons, even for tens of seconds, often leads to catastrophic error accumulation. And even when some world models do achieve long-horizon prediction ~\citep{wang2026interactive}, they often struggle with {\em generalization}, as they are unable to properly model the physics of the diverse contact-rich manipulation scenarios that robots commonly interact with.

We propose \modelname{}, a world model for real-time, long-horizon, contact-rich interaction across diverse objects and scenes. Our key contribution is a contact-based memory context, motivated by the need to remember how earlier interactions changed the scene even after objects left view. We combine a short window of recent frames with a set of long-term keyframes, capturing ongoing motion and past interactions without expanding the attention context. Inspired by human haptic memory \citep{hutmacher2018long}, we select keyframes at \emph{contact-making or breaking moments}, when robot--object interactions change the scene structure. The model predicts these events to update its memory during rollouts without external contact annotations, and retrieves keyframes by end-effector-pose similarity to recall relevant object states.

 Together with other architectual choices decisions detailed in the method section, we demonstrate that \modelname{} generalizes to \textbf{diverse scenes and tasks}, sustains \textbf{minutes of open-loop rollouts} under recorded actions, and \textbf{supports live teleoperation at 10 \,Hz}. Compared with state-of-the-art world models trained on the same robot dataset, DreamFAR produces more accurate predictions and accumulates less visual error over long rollouts. We evaluate this on 1,000 held-out scenes, where it achieves higher average prediction quality over one-minute rollouts (Fig.~\ref{fig:drift}). We also evaluate \modelname{} by the correlation between policy rollouts inside our model and the real world: across two policy checkpoints and three tasks, simulated and real-robot mean task progress have a Pearson correlation of $r=0.71$ (Fig.~\ref{fig:poleval}). Finally, we demonstrate that data collected inside the simulator can improve real-robot policy performance. Experimentally, training a policy on mixed real-world and simulator demonstrations increases the task progress by two-fold compared to a policy trained on real-world demonstrations alone.

\begin{figure}[t]\centering
\IfFileExists{figures/method.pdf}{\includegraphics[width=\linewidth]{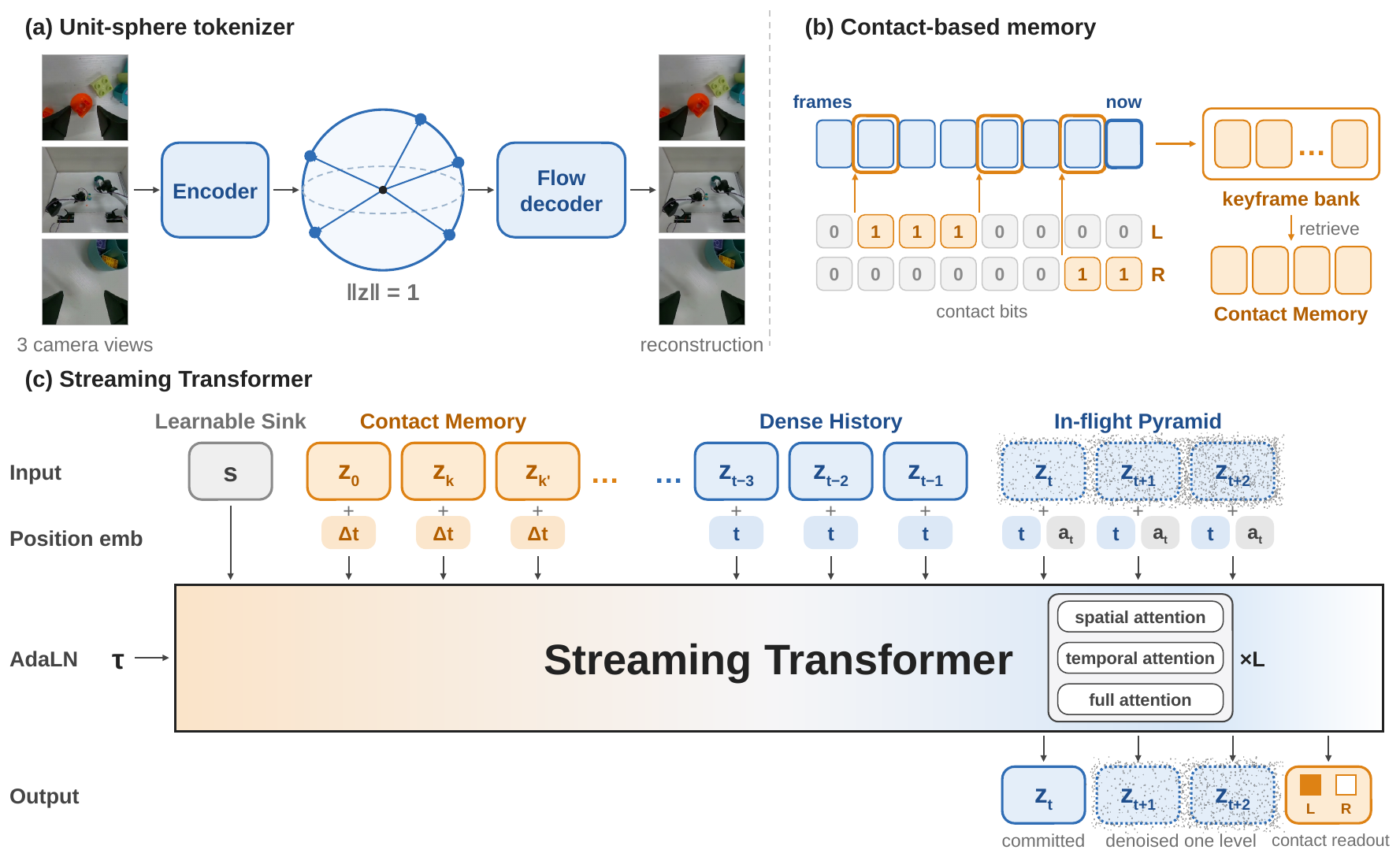}}{\fbox{\parbox[c][3.6cm][c]{0.97\linewidth}{\centering placeholder: (a) unit-sphere tokenizer, sequence [sink $|$ memory $|$ context], pyramid of in-flight frames, decoder; (b) contact bits, transitions, keyframes, pose-similarity retrieval into 8 slots }}}
\caption{\textbf{Overview of \modelname:} (a) Each view is encoded as per-position latents projected onto the unit sphere and decoded by a flow decoder. (b) When the robot makes or breaks contact with an object, we insert the corresponding keyframe into {\em contact memory}, which are index by end effector pose. (c) We use a transformer that has a number of design decisions that make it well suited to streaming and long-horizon generation. %
}\vspace{-2mm}\label{fig:method}
\end{figure}
\section{Related Work}\label{sec:related}

\paragraph{Video-generator world models.}
Most recent robot world models pretrain a large video generator on internet-scale videos, followed by post-training on robot specific data \citep{agarwal2025cosmos,ali2025cosmospredict,guo2025ctrl,ye2026dreamzero,gao2026dreamdojo}. These models typically contain billions of parameters, ranging from 2B to 14B \citep{agarwal2025cosmos,ali2025cosmospredict}. While large-scale video pretraining provides useful physical priors, accurately reproducing physical dynamics remains challenging \citep{kang2024physical,motamed2025physics}, and extending generation beyond the training horizon can lead to accumulating errors \citep{song2025history,zhang2025framepack}. \modelname{} takes the opposite route: its dynamics are trained causally from scratch on robot-native data and stream frame by frame.

\paragraph{Causal world models on robot or agent data.}
A parallel line of work train on robot-specific data from the start. Termed by \cite{ha2018world}, World Models jointly learn action-conditioned future observation prediction by factorizing the problem into representation learning (mapping high dimensional observations into latent state vectors), dynamics prediction (future state prediction conditioned on current state and actions) and future image generation (mapping a low dimensional future state into a high dimensional observation). Follow up works like DREAMER \citep{hafner2019learning} and many others use this framework for planning, RL and control \citep{hafner2025training,alonso2024diffusion,bruce2024genie}. Model-based RL \citep{silver2017mastering, deisenroth2011pilco, levine2013guided} works predate this recent interest in world modeling. On robots, action-conditioned video prediction from robot-only data dates to \citet{finn2016unsupervised} and continues with autoregressive, diffusion and feature-space predictors \citep{wu2024ivideogpt,zhu2024irasim,zhou2024dino,li2026causal}. Closest to ours are WEAVER \citep{jain2026weaver}, a causal world model trained on robot-native manipulation data, and the Interactive World Simulator (IWS) \citep{wang2026interactive}, which adds real-time, long-horizon interactive generation.

\vspace{-3mm}
\paragraph{Long-horizon and memory mechanisms.}
Two fundamental mechanisms allow modern video generators to model worlds; first, \emph{generation} which controls how errors compound, e.g., per-frame noise levels, KV-cached or rolling-window denoising, training on the model's own rollouts \citep{chen2024diffusion,yin2024slow,huang2026self,liu2025rolling}, and a frame-level attention sink on the initial generated frame(s) \citep{yang2025longlive,liu2025rolling}. Second, \emph{conditioning} which decides what to remember: compressed history, frame banks retrieved by camera pose or field-of-view overlap, or explicit 3D maps \citep{zhang2025framepack,xiao2025worldmem,yu2025context,wu2025video}; on robots, Ctrl-World and WEAVER keep pose-conditioned time-strided history frames \citep{guo2025ctrl,jain2026weaver}. \modelname{} adapts both to manipulation: the per-frame noise levels become a streaming pyramid of in-flight frames served from a rolling KV cache, the sink is one learned, never-noised, never-evicted frame, and memory keeps pose-similarity retrieval but changes what is eligible: keyframes are recorded when the contact transitions the model itself predicts (Sec.~\ref{method:memory}) happen.

\vspace{-3mm}
\paragraph{Policy evaluation and data generation.}
Physics simulators with real-to-sim assets rank policies in agreement with real trials \citep{li2024evaluating,zhang2025real,badithela2025reliable, jain2025polaris}, but each scene is built by hand. Learned world models skip the assets: policies are rolled out and scored inside the model \citep{quevedo2025worldgym,li2025worldeval,tseng2026sc3} or trained on generated data \citep{jang2025dreamgen}, and IWS reports real-to-simulated correlation of policy outcomes \citep{wang2026interactive}. \modelname{} targets the same two uses as a streaming simulator in which a policy, or a human through leader arms, acts closed-loop at 10\,Hz.

\section{Method}\label{sec:method}

Let $\mathbf{x}_t$ and $\mathbf{a}_t$ denote the visual observation and robot action at time $t$, and $h_t=(\mathbf{x}_{0:t},\mathbf{a}_{0:t-1})$ their history. We learn an action-conditioned world model that generates $K$ future observations autoregressively:
\begin{equation}
    p_\theta(\mathbf{x}_{t+1:t+K}\mid h_t,\mathbf{a}_{t:t+K-1})
    =\prod_{k=t}^{t+K-1}p_\theta(\mathbf{x}_{k+1}\mid h_k,\mathbf{a}_k).
    \label{eq:world_rollout}
\end{equation}
During rollout, $h_k$ incorporates the model's own predictions, enabling long-horizon simulation driven by recorded actions, a policy, or a human operator.

To achieve this goal, we propose \modelname, a latent flow-matching model that combines a visual tokenizer to encode observations into a compact latent space with a dynamics transformer to predict action-conditioned state transitions. We describe how unit-sphere normalization constrains latent scale (Sec.~\ref{method:tokenizer}), how predicted contact transitions and pose-based retrieval select long-term memories to complement recent context (Sec.~\ref{method:memory}), and how hybrid attention, pipelined denoising over a rolling KV cache, and a persistent attention sink enable streaming prediction (Sec.~\ref{method:dynamics}).

\subsection{Contact-Based Memory}
\label{method:memory}

A short context window can forget earlier interactions, such as where an object was placed after it leaves view. We therefore represent the visual history in $h_t$ using both a window of $T$ recent latents $\mathbf{z}_{t-T+1:t}$ and a long-term memory set $\mathcal{M}_t$. The recent window captures ongoing motion, while memory retains selected earlier observations without expanding the transformer's context with rollout length.

We select memory keyframes based on predicted contact transitions, which mark the onset or end of robot--object interactions. An auxiliary head on the dynamics transformer predicts two contact bits $\hat{\mathbf{c}}_t\in\{0,1\}^2$, one per arm; a change in either predicted contact adds the corresponding frame to the keyframe bank. The head is trained with binary cross-entropy loss against label predictions from a fine-tuned vision-language model \citep{clark2026molmo2}, with gradients detached from the dynamics backbone. During rollout, the model updates the bank using its own contact predictions, requiring no external contact labels.

To construct $\mathcal{M}_t$, we retain an initial anchor frame and retrieve keyframes preceding the recent window, ranked by end-effector-pose similarity to that window and recency. Poses are computed from predicted joint states via forward kinematics. Retrieved frames carry relative time offsets, and unused slots are masked, giving a fixed memory budget of eight frames. Labeling, retrieval, and training details are provided in App.~\ref{app:method}.

\subsection{Streaming Dynamics Transformer}\label{method:dynamics}

Our causal dynamics transformer predicts $\mathbf{z}_{t+1}$ conditioned on action $\mathbf{a}_t$, recent latents $\mathbf{z}_{t-T+1:t}$, and memory $\mathcal{M}_t$, while jointly predicting the robot's joint state, as well as contact. Hybrid attention combines spatial attention within frames and causal temporal attention across frames, with selected layers using joint spatio-temporal attention to capture motion across spatial positions. A learned sink frame~\citep{xiao2024efficient}, never noised or evicted, provides a persistent attention target even when memory slots are empty. We train with latent flow matching and independently sampled noise levels per frame, following diffusion forcing \citep{chen2024diffusion}.

For streaming inference, we maintain $P$ frames at staggered flow times and denoise them in parallel using a rolling KV cache \citep{liu2026rolling,jain2026weaver}. Each model call advances all in-flight frames, commits the oldest to the recent context, evicts the oldest context frame, and admits a new noise frame conditioned on the next action. Caching the committed context, memory, and sink keeps the dynamics computation per step constant in rollout length, while pipelining introduces a delay of $P$ steps between an action and its rendered frame.

\subsection{Unit-Sphere Tokenizer}\label{method:tokenizer}

Our tokenizer uses a finetuned SSDD~\citep{vallaeys2026ssdd} image encoder  to map each observation $\mathbf{x}_t$ to a compact latent $\mathbf{z}_t$ for dynamics modeling. We apply unit-sphere normalization \citep{davidson2018hyperspherical,loshchilov2024ngpt} to the encoded tokens and re-normalize predicted tokens before decoding and reuse as context to limit scale drift.

A flow-matching decoder $v_\phi$ reconstructs each camera view $\mathbf{x}_t^{(v)}$ conditioned on its latent $\mathbf{z}_t^{(v)}$. For Gaussian noise $\bm{\epsilon}\sim\mathcal{N}(\mathbf{0},\mathbf{I})$ and flow time $\tau\in[0,1]$, its objective is \citep{lipman2022flow,liu2022flow}
\begin{equation}
\begin{aligned}
    \mathbf{x}_{t,\tau}^{(v)} &= (1-\tau)\mathbf{x}_t^{(v)}+\tau\bm{\epsilon},  \mathcal{L}_{\mathrm{FM}}^{\mathrm{dec}}=\mathbb{E}_{\mathbf{x}_t,v,\tau,\bm{\epsilon}}\!\left[\left\|v_\phi(\mathbf{x}_{t,\tau}^{(v)},\tau,\mathbf{z}_t^{(v)})-(\bm{\epsilon}-\mathbf{x}_t^{(v)})\right\|_2^2\right].
\end{aligned}
\label{eq:decoder_flow}
\end{equation}
Here $t$ indexes observation time and $\tau$ indexes flow time; decoding integrates from noise at $\tau=1$ to an image at $\tau=0$. We jointly fine-tune the encoder and the decoder from a pretrained image tokenizer \citep{vallaeys2026ssdd}, and freeze them during dynamics training. Architecture and auxiliary training losses are detailed in App.~\ref{app:method}.

\providecommand{\experimentfigure}[2][]{%
  \IfFileExists{#2}{\includegraphics[#1]{#2}}{%
    \fbox{\parbox[c][2.4cm][c]{0.9\linewidth}{%
      \centering\footnotesize Figure asset not found:\\
      \texttt{\detokenize{#2}}}}}}
\ifdefined\ifexperimentstandalone\else\newif\ifexperimentstandalone\fi
\providecommand{\paperref}[2]{\ifexperimentstandalone #2\else\ref{#1}\fi}
\providecommand{\papercite}[2]{\ifexperimentstandalone #2\else\citep{#1}\fi}

\section{Experiments}\label{sec:experiments}
We first specify the experimental setup in \S\ref{sec:setup}. Next, we show the qualitative rollouts (\S\ref{sec:qual}) and the comparison against baseline methods (\S\ref{sec:baseline}) on long-horizon prediction. We apply \modelname{} to policy evaluation (\S\ref{sec:poleval}) and co-training with generated demonstrations (\S\ref{sec:cotrain}), and finally ablate the contact-based memory mechanism (\S\ref{sec:ablation}). 

\begin{figure}[t]
\centering
\experimentfigure[width=\linewidth]{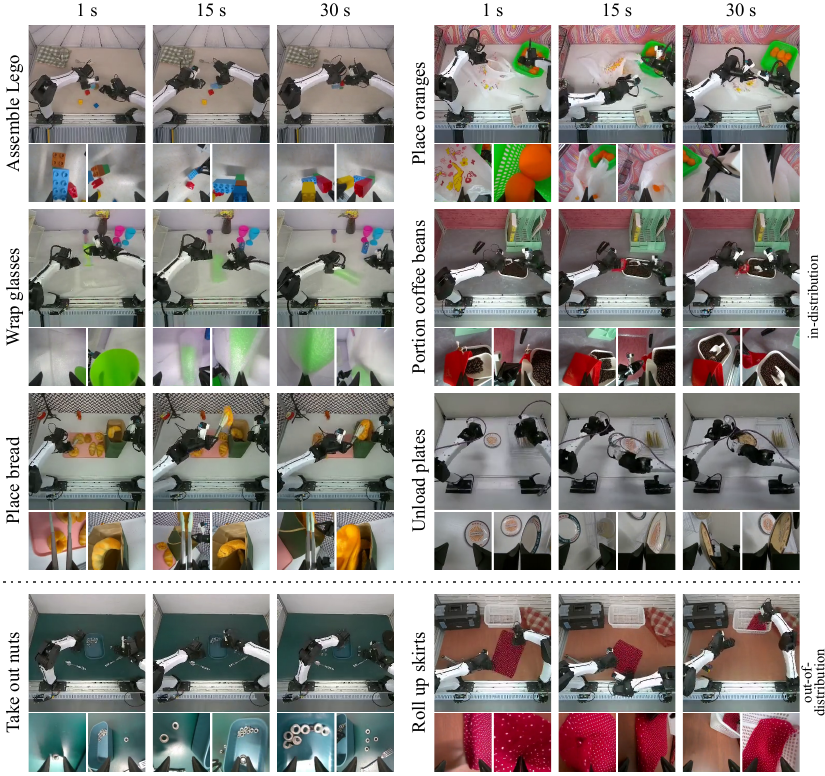}
\caption{\textbf{Rollouts on diverse interactions.} \modelname{} predicts accurate dynamics across diverse interactions, including compositional assembly (Lego), tool and container use (bags, bread, plates) and deformable objects (skirt).} 
\label{fig:objects}
\end{figure}

\vspace{-3mm}

\begin{figure}[t]
\centering
\includegraphics[width=\linewidth]{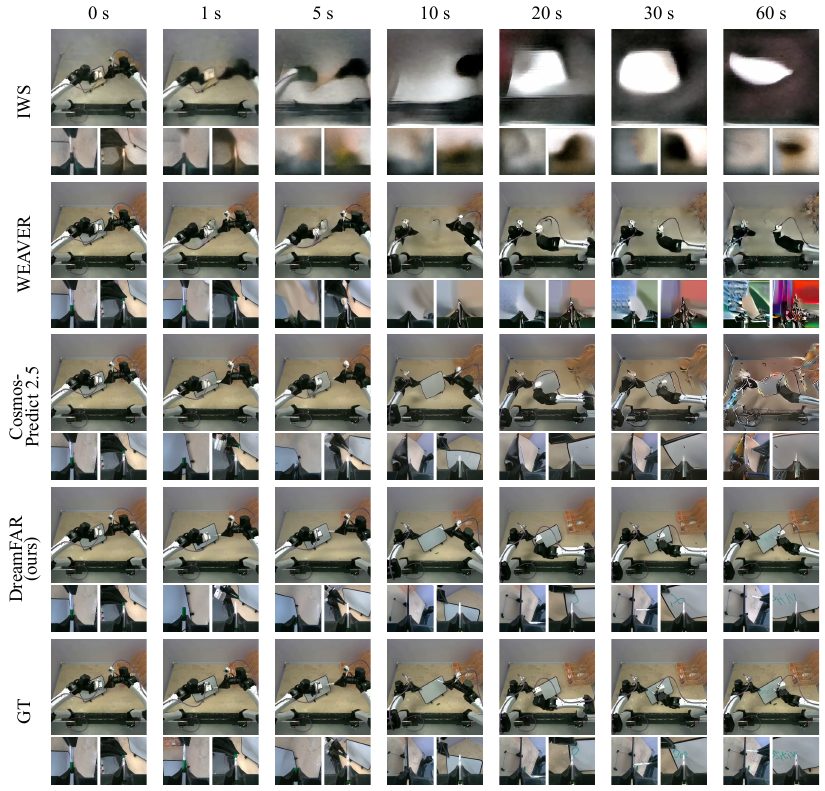}
\caption{\textbf{Visual comparison against baselines on a long-horizon prediction.} We run 60-s rollout on an episode of ``write a sentence on a whiteboard'' task, and qualitatively compare \modelname{} against the existing baseline models. IWS collapses within seconds, WEAVER hallucinates texture drastically in the wrist view, Cosmos-Predict~2.5 drifts and over-sharpens texture, while \modelname{} manages to keep the scene layout and the manipulated objects consistent over the whole minute, and it even predicts the writings on the whiteboard that is visually similar to the GT.}
\label{fig:qualcmp}
\end{figure}
\subsection{Experimental Setup}\label{sec:setup}

\textbf{Dataset.}
\modelname{} is trained on an internal bimanual teleoperation dataset collected on YAM robots: two 6-DoF arms with parallel grippers, a 14-D state and action space, and three camera views. 
We process all episodes to 10\,Hz and crop each of the three views to a square that is resized to $256\times256$. 
Actions are the 14-D joint targets at the same rate. 
For the baseline comparison, we instead train every model on the public ABC dataset~\citep{abc2026} for reproducibility: about 131k episodes over 201 tasks (roughly 3,500 hours). We hold out a fixed set of 1,000 ABC episodes for evaluation.

\textbf{Implementation details.}
Our tokenizer follows the architecture of SSDD-M \citep{vallaeys2026ssdd}, with a 46\,M-parameter convolutional encoder and a 50\,M-parameter decoder. Each $256\times256$ view becomes a $16\times16$ grid of 16-dimensional latents, with each position projected onto the unit sphere. Both are initialized from a pretrained SSDD checkpoint and fine-tuned jointly on our dataset. During fine-tuning, we add a uniformity~\citep{wang2020understanding} term to the loss function as a regularization.
The dynamics model of \modelname{} is a 1.9\,B-parameter transformer with 28 blocks of hidden size 2048 and 16 attention heads; every block applies spatial attention within a frame followed by causal temporal attention across frames, and five blocks use full spatio-temporal attention. Each frame is represented by 193 tokens (192 tokens for the visual frame and 1 token for the robot joint state). The model sees a 5-s window of 50 frames at 10\,Hz together with a fixed budget of eight memory slots and one learned sink frame.
Training is conducted on 256 H200 GPUs for six days with AdamW, a learning rate of $10^{-3}$ with 1,000 warm-up steps and no decay, a batch size of 2,048, and independently sampled logit-normal flow times per frame \citep{esser2024scaling,chen2024diffusion}. Inference uses $P=16$ frames in flight, a context-noise level of 0.24 on committed frames.

\vspace{-2mm}
\subsection{Qualitative Results}\label{sec:qual}
As is shown in Fig.~\ref{fig:objects}, \modelname{} produces stable rollouts across interaction types: compositional assembly of Lego bricks, tool use, and deformable objects. Objects keep their identity and position over the 30\,s, the arms follow the recorded actions, and contact events such as grasps and placements are rendered across the top and wrist views with high 3D consistency. 
The rollouts remain coherent in the OOD episodes, indicating that the learned dynamics transfer to unseen objects and task compositions.

\subsection{Baseline Comparison}\label{sec:baseline}\label{sec:abc}
We compare \modelname{} with three baselines: IWS \citep{wang2026interactive}, WEAVER \citep{jain2026weaver}, and Cosmos-Predict~2.5 \citep{ali2025cosmospredict}. 
For a fair comparison, we implement WEAVER and \modelname at a matched scale of about 0.8\,B parameters. We size IWS to about 0.4\,B to match the streaming step time. 
All models are trained on the same ABC split and evaluated on the 1,000 held-out episodes: each model is seeded with 15\,s of real frames and then rolled out autoregressively for 60\,s. We report PSNR, LPIPS, and FID in 10-s bins, averaged over episodes.

Fig.~\ref{fig:qualcmp} shows one held-out episode rollout over 60\,s. \modelname{} produces plausible frames throughout: the table layout, the arm pose and the manipulated object remain consistent with the real frames after 60\,s of rollout. Cosmos-Predict~2.5 progressively changes the arm geometry and the scene texture; WEAVER keeps the top view roughly coherent while the wrist views fill with hallucinated texture; and IWS collapses into a blurry image within the first 10\,s.

Fig.~\ref{fig:drift} confirms that \modelname{} drifts much more slowly than every baseline on all three metrics. Over the 60-s rollout, \modelname{} averages 16.4\,dB PSNR against 15.8 for Cosmos-Predict~2.5, 14.5 for WEAVER and 11.0 for IWS. 
The gap widens further on the perceptual and distributional metrics: \modelname{} has the lowest LPIPS in every bin and its FID rises from 15 to 32 over the minute, whereas Cosmos-Predict~2.5 rises from 30 to 73, WEAVER from 46 to 87 and IWS stays above 180 throughout.

\begin{figure}[t]
\centering
\includegraphics[width=\linewidth]{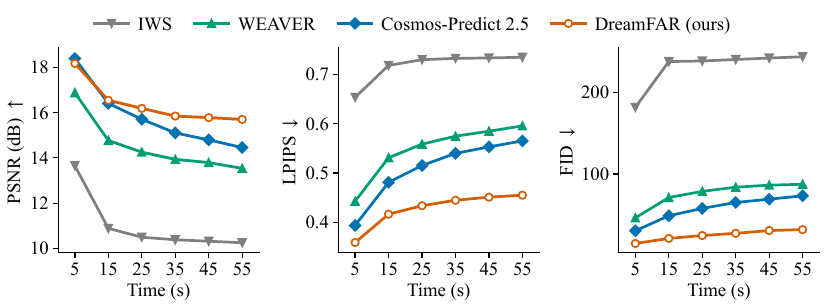}
\vspace{-3mm}
\caption{\textbf{Metric Drift on ABC over one minute.} We evaluate the video prediction performance of \modelname{} and the existing baseline models on standard video quality metrics, including PSNR, LPIPS and FID. We run 60-s rollout (600 frames) on 1,000 held-out episodes and average the results into 10-s bins. \modelname{} degrades the slowest on all three metrics and significantly outperforms all baseline models, indicating the stability of its long-horizon prediction.}
\label{fig:drift}
\end{figure}
\vspace{-3mm}

\subsection{Policy Evaluation}\label{sec:poleval}

\begin{figure}[t]
\centering
\begin{minipage}[b]{0.545\linewidth}\centering
\experimentfigure[width=\linewidth]{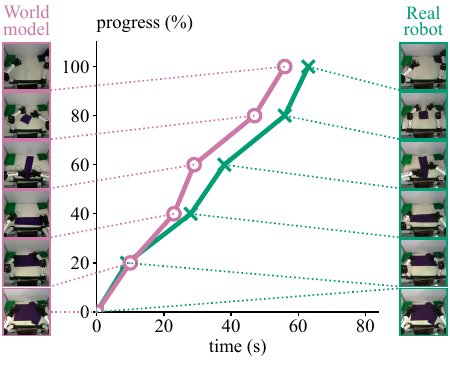}\\[-1pt]{\small (a)}
\end{minipage}\hfill
\begin{minipage}[b]{0.44\linewidth}\centering
\experimentfigure[width=\linewidth]{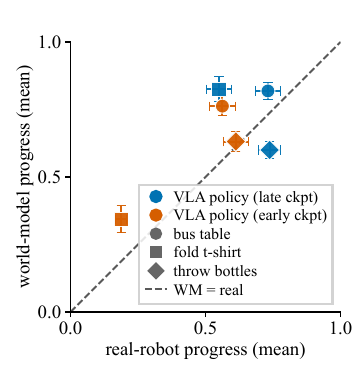}\\[-1pt]{\small (b)}
\end{minipage}
\caption{\textbf{Policy evaluation inside \modelname{} against the real robot.} We run policy evaluation on both the real robot and inside the \modelname{}. (a) shows the task progress over time for one episode of the ``t-shirt folding'' task, on the real robot ($\times$) and inside \modelname{} ($\circ$). (b) summarizes the mean progress per policy and task. The simulator is generally optimistic in absolute progress but preserves the ranking on two of three tasks.}
\label{fig:poleval}
\end{figure}

We evaluate whether \modelname{} can serve as a faithful proxy for real-robot policy evaluation. We define three tasks: placing plates into a bin, throwing plastic bottles into a bin, and folding a t-shirt. 
The policy is $\pi_0$ \citep{black2024pi_0} fine-tuned on teleoperation demonstrations of our tasks, and we compare its late checkpoint (15k steps) against an early checkpoint (2.5k steps). 
For every policy--task pair we run 32 episodes on both real robot and inside \modelname{}, with the same prompt; each episode is hand-labelled with a task progress in $[0,1]$, the fraction of the task completed within the episode.
Fig.~\ref{fig:poleval} summarizes the results. Per-task mean progress in the simulator correlates with the real robot at Pearson $r=0.71$ over the six policy--task points. The simulator is optimistic in absolute terms: it tends to report higher progress score than the real robot, and in the episode of Fig.~\ref{fig:poleval}(a) the simulated rollout reaches every stage of the fold 5--10\,s before the real one.

\subsection{Co-Training with Generated Demonstrations}\label{sec:cotrain}

\begin{figure}[t]
\centering
\begin{minipage}[c]{0.42\linewidth}\centering
\experimentfigure[width=\linewidth]{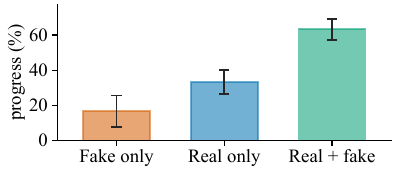}
\end{minipage}\hfill
\begin{minipage}[c]{0.55\linewidth}
\caption{\textbf{Co-training with demonstrations collected inside \modelname{}.}
\modelname{} enables real-time teleoperation without a real robot, and we collect 75 demonstrations of the ``place bags of chips into a basket'' task in both real and simulated environments. The $\pi_0$ policy finetuned on the combination of both nearly doubles the real-only performance.}
\label{fig:cotrain}
\end{minipage}
\end{figure}

Since \modelname{} runs in real time from teleoperation input (Fig.~\ref{fig:teaser}), an operator can collect demonstrations inside the simulator without a real robot: the GELLO leader arms \citep{wu2024gello} drive the simulated robot at 10\,Hz, and each episode is recorded in the training-data format, with the rendered views, the predicted joint state and the commanded actions. 
We test whether such simulated demonstrations are useful for policy learning on a single task, placing three bags of chips into a basket. We fine-tune the $\pi_0$ policy \citep{black2024pi_0} on three datasets: 75 real demonstrations, 75 simulated demonstrations, or their union (150 episodes). 
Each policy is evaluated in 10 real-robot trials of 60\,s.
Fig.~\ref{fig:cotrain} reports the results. Combining the two nearly doubles the real-only result to $63\%$. This study shows that demonstrations collected inside \modelname{} carry useful signal for the real robot, rather than isolating the value of a simulated episode against a real one.

\subsection{Ablation on Contact-Based Memory}\label{sec:ablation}

\begin{figure}[t]
\centering
\experimentfigure[width=\linewidth]{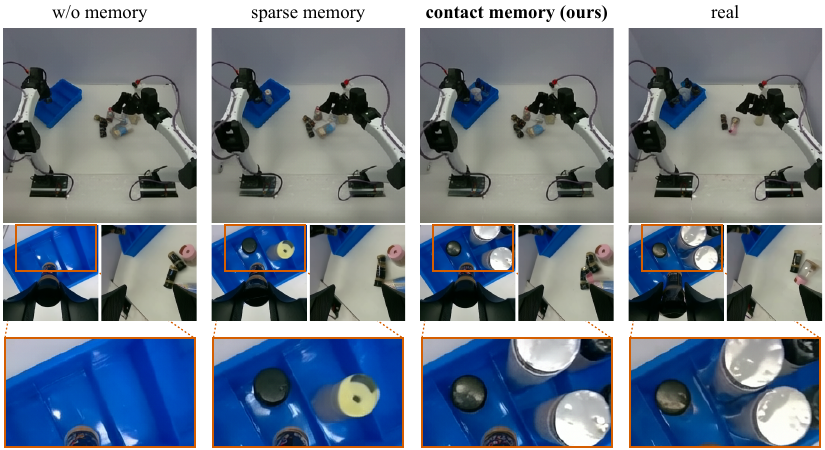}
\vspace{-3mm}
\caption{\textbf{Contact-based memory keeps placed objects persistent.}
Three models with the same architecture but different memory selection. We show one frame from a rollout of each model. Without memory the bin is empty, sparse time-strided memory hallucinates its contents, and contact-based memory matches the real observation.}
\label{fig:memory_ablation}
\end{figure}
\begin{table}[t]
\centering
\begin{minipage}[c]{0.47\linewidth}\centering
\small
\resizebox{\linewidth}{!}{%
\begin{tabular}{lccc}
\toprule
Metric & No memory & Fixed stride & Contact (ours) \\
\midrule
PSNR$\uparrow$ & 18.655 & 18.988 & \textbf{19.251} \\
LPIPS$\downarrow$ & 0.348 & 0.334 & \textbf{0.326} \\
FID$\downarrow$ & 37.156 & 36.438 & \textbf{34.811} \\
DINO$\uparrow$ & 0.749 & 0.767 & \textbf{0.777} \\
\bottomrule
\end{tabular}}
\end{minipage}\hfill
\begin{minipage}[c]{0.50\linewidth}
\caption{\textbf{Contact-based memory improves long-horizon prediction.}  We run rollout of up to 60\,s using the three models of Fig.~\ref{fig:memory_ablation} on the held-out episodes. Metrics against the real video are averaged over the two wrist views. DINO is the cosine similarity of DINOv2 features \citep{oquab2023dinov2}. Contact-based memory achieves the best performance on every metric.}
\label{tab:memory_ablation}
\end{minipage}
\end{table}
\vspace{-3mm}

We ablate the memory design of Sec.~\ref{method:memory} on a task with 4.6k episodes where objects leave and re-enter view: ``organize condiment bottles''. Three models differ only in memory: no memory, sparse memory with keyframes taken at a fixed 1-s stride, and our contact-based memory. 
To isolate the effect, all three use a short 1-s recent context and two memory slots.
Fig.~\ref{fig:memory_ablation} shows the three rollouts and the real observation when the left wrist camera leaves and looks into the bin again. Without memory, the bin is rendered empty: the bottles placed earlier are no longer in the context and the model produces an unoccupied bin. 
With sparse memory, the bin contains bottles, but the wrong ones. With contact-based memory, the bin contents match the real observation. This comes from the mechanism: a predicted contact transition, here the release of a bottle, writes exactly the frame that shows the new scene state into the keyframe bank, and when the end-effector returns to a similar pose the frame is retrieved as clean context. The memory therefore makes placed objects persistent across the rollout at a fixed cost in context length, enabling long rollouts of Secs.~\ref{sec:qual} and ~\ref{sec:baseline}.
Tab.~\ref{tab:memory_ablation} quantifies the effect. We compare the predictions with the real video on the two wrist views using PSNR, LPIPS, and FID. Contact-based memory is best on every metric, indicating the impact of our design choices.

\vspace{-3mm}
\section{Discussion and Limitations}\label{sec:discussion}
\vspace{-3mm}

We presented \modelname{}, a world model for real-time, long-horizon, contact-rich manipulation across diverse objects and tasks. By combining long--short-term memory, keyframe selection based on predicted contact transitions, and streaming dynamics with a persistent attention sink, our framework supports extended interactive rollouts and applications in policy evaluation and data generation.

\textbf{Scene diversity.} Our model is limited to interactions within a YAM station. Future work can extend this framework to open-world environments with more diverse backgrounds, layouts, and scenes.

\textbf{Robot embodiment.} Our model captures a single robot embodiment. Learning a world model across multiple embodiments is a next step toward simulating interactions across robots with different kinematics and action spaces.

\textbf{Manipulation dexterity.} The YAM station's grippers offer less dexterity than multi-fingered hands, limiting the range of manipulation behaviors studied here. Future work can extend our model to more dexterous robots and fine-grained manipulation tasks, such as in-hand object reorientation.

\section*{Ethics Statement}
\modelname{} is trained on recordings of robot teleoperation: camera views of a bimanual robot workspace together with the robot's joint states and commanded actions. The recordings contain no personal data, no human subjects were studied, and no identity, biometric or demographic information is collected or used; the teleoperators are not identified and appear in no released artefact. The public ABC dataset is used under its Apache-2.0 licence and cited by its paper. The simulator is trained on predominantly successful demonstrations and may over-estimate the success of a policy; it must not be the sole gate for deploying a policy, and policies evaluated inside it must still be validated on hardware. Training the model required a substantial multi-GPU compute budget; releasing the weights avoids repeating it.

\section*{Reproducibility Statement}
The architecture, training recipe and inference procedure are described in Sec.~\ref{sec:method}; the evaluation protocols, baseline adaptations and inference settings are described in Sec.~\ref{sec:experiments}. We release the code for training, inference and the interactive demo, and the model weights of the full \modelname{} and of the ABC-trained \modelname{}. With the release, the equal-data comparison and the speed measurements can be reproduced: we release our ABC split (training tasks, held-out episodes and the 10 held-out tasks), the benchmark episode list, the square-crop preprocessing and the evaluation scripts.

\section*{AI Use Statement}

In this work, we used generative AI tools to implement methods (writing and refactoring parts of the training, inference and evaluation code) and to support the interpretation of results. We have not used generative AI tools to generate
synthetic data sets, develop theoretical models or conceptual frameworks, formulate or prove mathematical claims, propose
or refine hypotheses, or design the research methodology or experiments; translation, dataset cleaning and qualitative or
thematic data analysis are not applicable to this work. 
Additionally, we used generative AI tools to edit parts
of the paper for readability. 
We have reviewed all AI-assisted work: LLM-generated code was verified and tested for correctness
by the authors, and all claims, numbers, figures and citations were checked by the authors against our experiment records
and the cited sources. 
We take responsibility for the final content of this work, including text, claims or artifacts
produced with the aid of generative AI.

\bibliographystyle{plainnat}
\bibliography{iclr2027_conference}

\appendix
\section{Architecture and Training Details}\label{app:method}

\paragraph{Tokenizer recipe.} Each observation $\mathbf{x}_t$ contains three $256\times256$ RGB views. A shared convolutional encoder with stride 16 (46.1\,M parameters) maps each view to a $16\times16$ grid of 16-dimensional vectors; $\mathbf{z}_t$ collects the latent tokens from all three views. The decoder is a U-ViT flow-matching decoder (50.3\,M parameters) \citep{bao2023all}. The encoder output is L2-normalised per spatial position; there is no KL term, and one learned scalar inside the decoder rescales the latent, so the served latent is exactly unit-norm (measured mean norm 1.000, standard deviation $10^{-4}$). The decoder reconstructs each view from noise conditioned on its latent tokens. Its loss is the flow-matching error \citep{lipman2022flow,liu2022flow} plus LPIPS (weight 0.5) \citep{zhang2018unreasonable}, representation alignment (0.1) against a frozen vision-transformer teacher with a warm-up \citep{yu2024representation,fu2026vision}, and a uniformity term (0.1), a hinged squared-MMD reformulation of the loss of \citet{wang2020understanding}. Decoding takes two Euler steps, about 14\,ms per view-frame; for video, one noise tensor is drawn per episode and reused for every frame. Tab.~\ref{tab:tok_recon} reports reconstruction on held-out episodes; on held-out tasks PSNR is 26.51 / 30.06 / 30.65\,dB for the top / left-wrist / right-wrist views.

\begin{table}[h]
\centering
\caption{\textbf{Tokenizer reconstruction on held-out episodes.} Single-frame reconstruction at $256\times256$ per view; PSNR in dB.}
\label{tab:tok_recon}
\begin{tabular}{lccc}
\toprule
View & PSNR $\uparrow$ & SSIM $\uparrow$ & LPIPS $\downarrow$ \\
\midrule
Top & 27.95 & 0.722 & 0.085 \\
Left wrist & 31.26 & 0.664 & 0.100 \\
Right wrist & 31.55 & 0.670 & 0.100 \\
\bottomrule
\end{tabular}
\end{table}

\paragraph{Sequence layout and positions.} The three $16\times16\times16$ view latents are stacked into a $48\times16\times16$ tensor per frame and patchified with patch size 2 (192 tokens) plus a state token carrying the 14-D joint positions, 193 tokens per frame. The backbone has 28 blocks with hidden size 2048 and 16 heads of width 128. Spatial attention uses 2-D rotary positions and temporal attention rotary positions on the frame index; the sink frame (193 learned vectors) receives no rotation, memory frames are rotated by a log-compressed time offset $\Delta t$ to the current frame, never by a slot index, and unused slots are masked. The five joint-attention blocks use a frame-causal block-sparse mask. Every attention applies qk-RMSNorm \citep{dehghani2023scaling}; the flow time enters through adaptive layer normalisation with zero initialisation \citep{peebles2023scalable}, and each frame's action conditioning is added to that frame's modulation through a small MLP.

\paragraph{Dynamics training.} The flow loss is applied on the context positions with independent logit-normal flow times per frame \citep{esser2024scaling,chen2024diffusion}; the state token carries a flow loss with weight 0.1 and the contact readout a BCE loss with weight 0.1. Memory slots are dropped with probability 0.2. We use AdamW with learning rate $10^{-3}$, 1k warm-up steps and then a constant schedule, $\beta=(0.9,0.95)$, weight decay 0.01, gradient clipping at 0.5 and an EMA of the weights with decay 0.999; the EMA weights are used at inference. The effective batch size is 2048 windows. The model was initialised, weights only, from an earlier training run of the same architecture taken at 14.5k of its steps.

\paragraph{Contact labeler.} Each call of the fine-tuned vision-language model \citep{clark2026molmo2} receives, for one arm, an 8-frame clip at 5\,Hz of composites of that arm's wrist view and the base view, and returns per-frame booleans with a soft probability from the token logits; two passes offset by one frame are interleaved to 10\,Hz and smoothed with a 3-frame box filter, a hysteresis (enter above 0.6, leave below 0.4) and a minimum dwell. The model was fine-tuned on human-corrected labels with task-disjoint splits. In a visual audit, contact onsets lag gripper closing by 0.03--0.38\,s, releases coincide with re-opening, and sustained holds stay at high probability with no gripper motion, so the detector is not a gripper proxy. A re-run reproduces the reference predictions at 99.2\% per-frame agreement, a determinism check rather than a validity measure. Rejected design choices: a gripper open/close hysteresis as labeler (blind to non-prehensile contact; misses about 60\% of grasps on the achieved gripper state), field-of-view-overlap retrieval keys (degenerate under a static top camera), adaptive contact thresholds (they leak future statistics), and co-denoising contact as a flow modality (the detached readout was better calibrated).

\end{document}